# Space-Filling Curve Indices as Acceleration Structure for Exemplar-Based Inpainting

Tim Dahmen[1,3,*], Patrick Trampert[1,2,3], Pascal Peter[2], Pinak Bheed[2], Joachim Weickert[2] and Philipp Slusallek[1,2]

*[1] German Research Center for Artificial Intelligence (DFKI) GmbH, Saarbrücken, Germany*
*[2] Saarland University, Saarbrücken, Germany*
*[3] Tim Dahmen and Patrick Trampert contributed equally to this manuscript*
** corresponding author: Tim.Dahmen@dfki.de*

***Abstract* —Exemplar-based inpainting is the process of reconstructing missing parts of an image by searching the remaining data for patches that fit seamlessly. The image is completed to a plausible-looking solution by repeatedly inserting the patch that is the best match according to some cost function. We present an acceleration structure that uses a multi-index scheme to accelerate this search procedure drastically, particularly in the case of very large datasets. The index scheme uses ideas such as dimensionality reduction and *k*-nearest neighbor search on space-filling curves. Our method has a theoretic runtime of $O(\log^2 n)$ per iteration, where n is the size of the dictionary. It reaches a speedup factor of up to 236 over the original brute-force method. The approach has the advantage of being agnostic to most model-based parts of exemplar-based inpainting such as the order in which patches are processed and the cost function used to determine patch similarity. Thus, the acceleration structure can be used in conjunction with most exemplar-based inpainting algorithms.**

## I. Introduction

The term "inpainting" refers to the reconstruction of missing parts of an image. The technique is typically applied to two-dimensional images such as photos or digital micrographs, but can also be applied to three-dimensional datasets such as videos or tomograms [1]. The most common applications for inpainting are the restoration of damaged photos and the synthetic filling of holes in images where unwanted parts have been deleted. The basic principle is quite versatile and can be applied to less common domains as well. In inpainting-based super resolution algorithms, the resolution of an image is enhanced a-posteriori by synthetically generating additional pixels using inpainting methods [2]. In sparse scanning microscopy, only a subset of the pixels of a digital micrograph is acquired and missing parts of the image are generated synthetically using inpainting algorithms [3].

Most relevant to this work are algorithms from the class of exemplar-based inpainting [4–6]. Exemplar-based inpainting works iteratively on small, often rectangular (or cubic in the three-dimensional case) parts of the image, called patches. In a first step, a target patch is selected for processing. The target patch is always selected at the border of the missing region of the image, such that it overlaps both the missing region of the image and a region of the image where pixel values are known. The selection is done heuristically, whereby the applied heuristic has a crucial impact on the quality of the inpainting result. The work of Criminisi et al. uses a heuristic that aims to continue linear structures [5]. This approach has been extended to also detect incomplete salient structures, which incorporates curvature [7]. These structures are inpainted first to guarantee that the most important edges of an image are connected correctly. The remaining sub-regions are then inpainted using the approach from Criminisi. Instead of detecting structures automatically, a user can also interactively specify salient structures as input [8]. Applying belief propagation, specified structures are reconstructed first, followed by the remaining missing parts. The concept of structure sparsity of a patch was introduced in [9]. Structure sparsity is a measure for the sparseness of the nonzero similarities to neighboring patches. Patches with higher structure sparseness are filled first. In [10], an inpainting order is generated by interpolating level lines with Euler spirals. In [11] a filling order based on color values was proposed.

Once a target patch has been selected, the algorithm searches known parts of the image, called the dictionary, for a suitable patch to be inserted into the missing area. The selection is performed using a cost function, for example the $L_2$-norm between known pixels of the target patch and corresponding pixels of a patch in the dictionary. The cost function is another degree of freedom that allows tailoring an inpainting algorithm to different situations. The $L_2$-norm can be weighted with the Hellinger distance [12], replaced with the structural similarity index [13], or converted so that it is equivalent to a Lagrangian relaxation of a strictly guided reconstruction [10]. Rather than inpainting individual patches, a linear combination of several most similar patches, whose coefficients are computed by a constrained optimization problem, can be used [6]. A completely different approach is a variational framework based on partial differential equations to address exemplar-based inpainting [14]. Ibarrola et al. [15] proposed a two-step procedure: First, curvature-driven diffusion inpainting is used to build a pilot image that is used to infer a-priori structural information on the image gradient. Second, the image is inpainted by minimizing a mixed spatially variant anisotropic functional that uses the pilot image.

No matter which cost function is used, the algorithm always selects the patch from the dictionary that evaluates to the global minimum of the cost function value relative to the selected target patch. Typically, this selection is performed by means of a full scan, i.e. the cost function is evaluated for every patch in the dictionary and the patch corresponding to the minimum cost function value is selected. A dictionary can be huge with about $10^6$ entries in the two-dimensional case and more than $10^8$ in the three-dimensional case, such that the full scan becomes computationally very expensive and is the performance limiting factor of the algorithm.

Various solutions have been proposed to improve the performance of exemplar-based inpainting, including probabilistic methods and local search strategies. Sangeetha et al. [16] have proposed a cost function that includes a diameter of the region surrounding the patch to be inpainted. Depending on this diameter, only the local area around the patch is taken into consideration for finding a patch to be inserted. Ruzic et al. [17], [18] used contextual features to guide the search for image patches. The image is divided into non-overlapping blocks that are used to reduce the search space for each patch to be inpainted, which also reduces the computation time. Alilou et al. [19] introduced a new search strategy that does not need to calculate the distance of all pixels contained in a patch. Combined with an early stopping criterion this reduced computation time by a factor of two to five. Kwok et al. [20] proposed to decompose patches into frequency coefficients. Using only the most significant coefficients to compute matching scores, they were able to reduce computation time by a factor of around 11. He et al. [21] used offset statistics to get a subset of the source region as optimal search area for a patch to be inserted. Their method performs two to five times faster than a full scan. Barnes et al. [22] have proposed PatchMatch as a generic acceleration strategy for exemplar-based algorithms. It relies on a combination of propagating knowledge of already found patch pairs and interleaved random searches to avoid local minima in the patch optimization process. Among other image editing algorithms, they have also proposed inpainting as an application. In particular, they have incorporated PatchMatch into the inpainting algorithm of Wexler et al. [23].

Our work was influenced by several other concepts. The idea to consider patches as data points in a high-dimensional space was first developed by Buyssens et. al. [24]. However, the work does not apply techniques from high-dimensional point search but uses a search-window-based strategy. The concept to order patches in a one-dimensional array according to some similarity-measure was first mentioned in [25], but used for the purpose of filtering, not searching. The idea of building an index structure over such data points was first proposed in [26]. However, the work used tree structured vector quantization as the search algorithm and it was applied to texture synthesis, not inpainting.

### *A. Our Contribution*

We propose to use high-dimensional indexing techniques as typically applied in the field of multimedia databases as acceleration structures to the patch search. We show how the problem of searching the best patch for a target location corresponds to a nearest neighbor search in an ad-hoc subspace of the dictionary. As the most general form of this problem is suspected to be linearly hard, we use additional knowledge on the inpainting algorithm to map the patch search on the much simpler problem of a nearest neighbor search in a subspace that is approximately known a-priori. This problem can then be addressed using techniques that are well known from the field of high-dimensional indexing, such as dimensionality reduction by means of a principal component analysis, filter-and-refine approaches using k-nearest neighbor (knn) queries, and multi-dimensional indexing using space-filling curves. Furthermore, we adapted PatchMatch to work with exemplar-based inpainting algorithms that insert whole patches following some priority measure and compared it against our proposed acceleration structure. Our solution has the advantage that it is independent of most model-specific aspects of exemplar-based inpainting, so that it is applicable for a large variety of inpainting algorithms.

# II. Material and Methods

Our inpainting process starts by building the acceleration structures called indices. The basic idea is to split the dictionary into patches of constant size and arrange them in a one-dimensional array, sorted by some measure of similarity, such that similar patches are stored close to each other. The indices are built over patches of constant size K×K. For colored RGB images, red, green and blue values are treated separately, resulting in $K^2 \cdot 3$ values per patch. Typical patch sizes are $K$=9 to $K$=13, so a patch corresponds to at least 9·9·3=243 values. In the following, we consider a patch as a point in a $K^2$-dimensional (or $K^2 \cdot 3$, respectively) space, called the patch space.

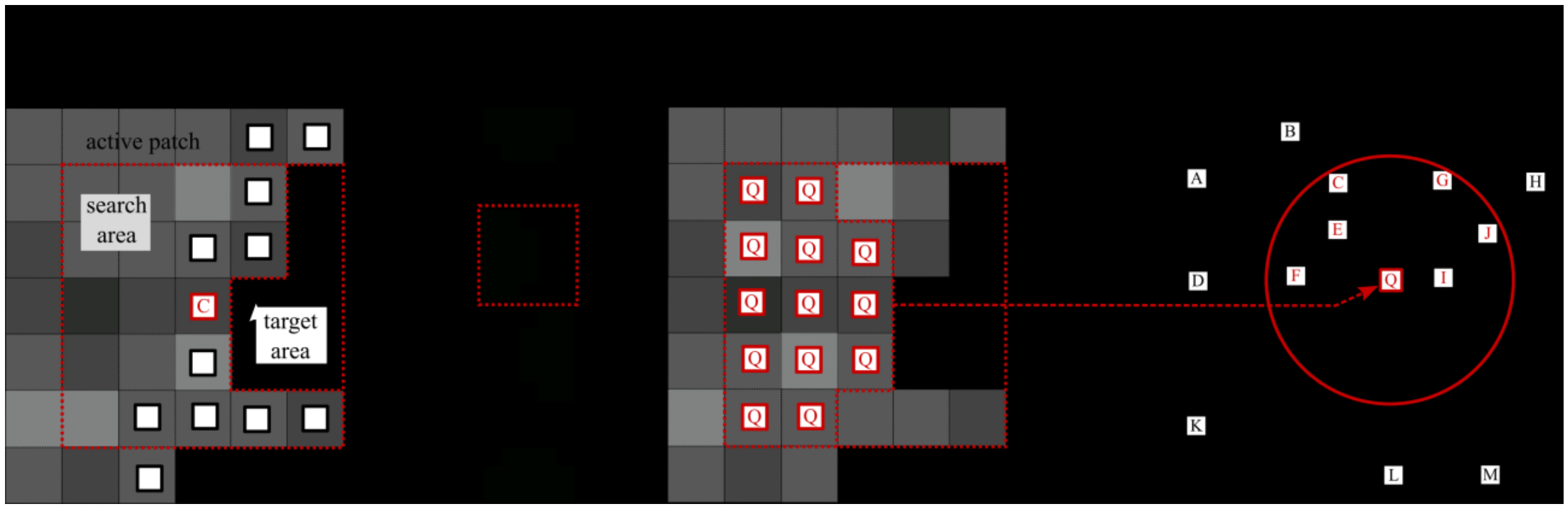

Figure 1. Overview of the patch search algorithm. a) The center of the active patch (marked C) is selected on the fillfront (white squares). The fillfront is defined as pixels with known value adjacent to pixels with unknown value. The known pixels inside the active patch are called search area, the unknown pixels are called target area. b) Index 2 is selected as it best covers the search area. c) The query object is formed by all pixels covered by this index, marked Q. It is by definition a subset of the search area. d) The query object is transformed to principal space and a $k$-nearest neighbor search is performed to select $k$ candidate patches. The best patch is then selected by means of a full scan in image space on only those $k$ candidates.

During the inpainting procedure, some values of the processed patch are unknown, as they are about to be inpainted. Patch similarity is therefore always defined on a subset of pixels, and the cost function must be defined in such a way that it can handle unknown pixel values. In the example of the $L_2$-norm, one simply assumes that the difference between unknown pixel values is zero. Geometrically, this can be interpreted as a parallel projection to a lower-dimensional subspace. If a pixel value is unknown, the dataset is projected parallel to the dimension that corresponds to the unknown pixel value, and distance is measured in the resulting, lower-dimensional space. As different pixels are known in every inpainting step, we say that we measure patch similarity in an ad-hoc subspace of the patch space.

We build the indices using a three-stage technique. First, we select a subset of the pixels to consider in each index. Second, we reduce the dimensionality of the resulting data points by means of a principal component analysis. Finally, we arrange them on a one-dimensional array according to their position on a space-filling curve (Figure 1).

Table 1. Annotations used throughout this paper.

| symbol | meaning | typical values |
|---|---|---|
| $K$ | patch size | 7 to 13 |
| $c$ | fraction of pixels used in index | 0.6 |
| $D$ | dimensionality of principal space | 6 to 12 |
| $k$ | size of candidate set (filter-and-refine) | 40 to 160 |
| $\mu$ | minimal interval size to use recursion | 128 to 512 |
| $v$ | minimal interval size for parallelization | 512 to 2048 |

In order to approximate the ad-hoc subspaces, we build several indices, where each index covers a different subset of the pixels in the patch. Each index uses a fraction $c$, typically about 60%, of these values (see the result section for a determination of optimal values of $c$). For inpainting two-dimensional images, we build eight separate indices, where each index contains the pixels nearest to the middle of one of the edges or one of the corners of the patch (Figure 2).

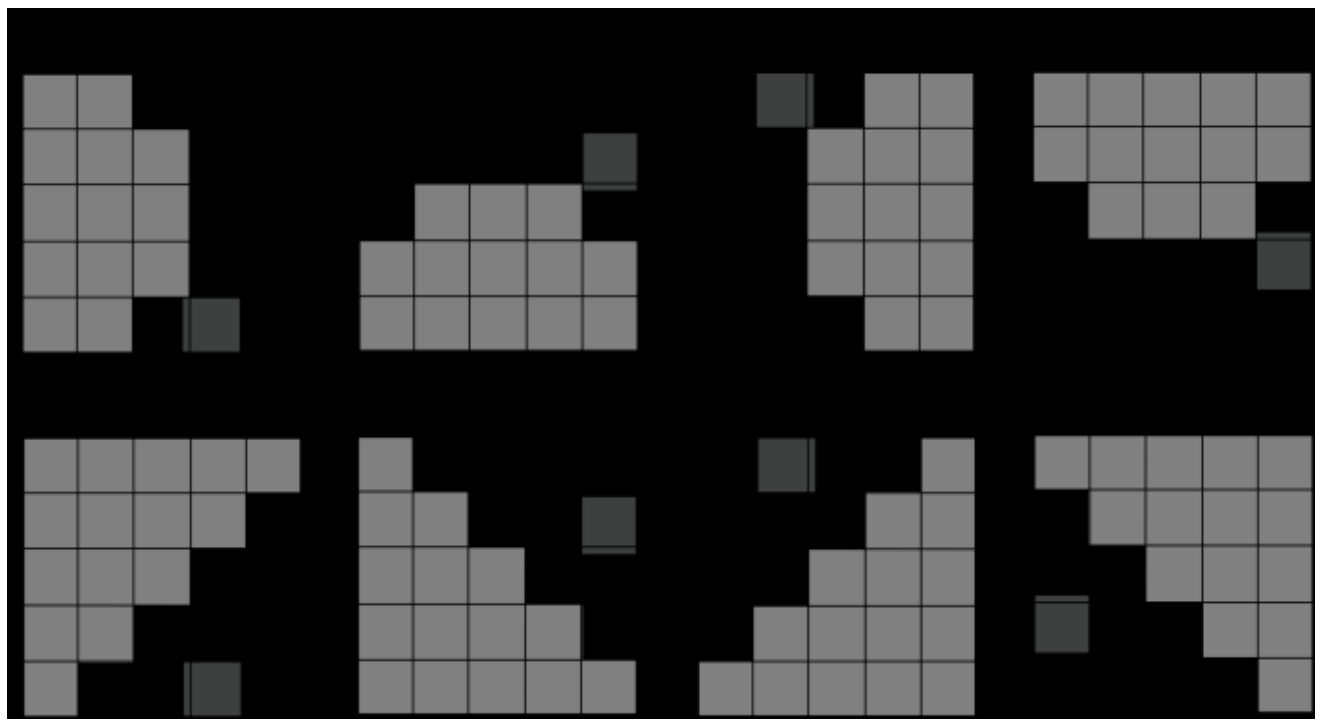

Figure 2. We build a total of eight indices. Each index uses approximately 60% of the pixel values of the patch. (a-d) Four indices use pixel values closest to the middle of a patch edge. (e-h) The remaining four indices use pixel values closest to a corner of the patch.

### *A. Dimensionality Reduction*

All known index structures with support for efficient nearest neighbor queries suffer from the curse of dimensionality, which means they scale well only for a total number of 8 to 16 dimensions. In the field of high-dimensional indexing, techniques of dimensionality reduction are therefore used prior to building the index. Dimensionality reduction can be performed by several techniques. Common approaches are principal component analysis (PCA) and linear discriminant analysis. Recently, a technique called random projections (RP) [27] has also received widespread interest, as it is computationally cheaper than a PCA, particularly in high-dimensional spaces. However, we implemented and tested RP and found that RP does not preserve distances sufficiently well for our application. Therefore, we reduce the dimensions of the index to a constant number $D$ by using a conventional PCA of which the first $D$ principal components are kept for further processing.

### *B. High-Dimensional Indexing*

After dimensionality reduction, an index can be built based on a moderate number of dimensions. A large number of index structures has been proposed for this purpose, including the well-known R-tree [28] plus its variants $R^+$-tree [29] and $R^*$-tree [30]. Slightly better scalability up to 20 dimensions was reported for the structure's X-tree [31], SS-tree [32], TV-tree [33], and the proxy-based indexing method called iDistance [34]. More recently, the idea of locality preserving hashing [35] has also gained some interest. However, most of the work on multi-dimensional indexing comes from the field of database design and considers out-of-core situations in the sense that the data is stored in form of fixed-size pages. An index should optimize the number of page accesses in this case. In our situation, several indices are stored in main memory, such that the memory consumption per index as well as the computational cost for traversing the index become relevant factors. Memory locality is still an important factor as cache hit rates have a high impact on CPU performance, but to a much lesser extent than for disc-based access schemes. Therefore, we selected an index structure that has low memory consumption per element and supports computationally efficient $k$-NN-queries.

The basic idea is to use a space-filling curve (SFC) to map the dictionary space to a one-dimensional sequence. A large number of SFCs has been investigated in the past, with early work including the famous Hilbert SFC [36], Peano SFC [37], and Lebesgue SFC [38]. A good introduction to the topic of SFCs is given in [39]. Of most practical relevance are z-curves [40], which can be considered to be polygonal approximations of the Lebesgue SFC on a discrete space. Z-curves have the distinct advantage that the computation of positions on the curve is computationally very cheap. In our approach, we use a z-curve to map the dictionary space to a one-dimensional structure. This structure is sorted and can be used by a binary search to find points in the set efficiently.

### *C. Sorting Relative to Z-Curve Position*

After performing a PCA and transferring all patches in the dictionary to principal space, a vector with the patch positions must be sorted according to the position of the principal space representation of the vector on the z-curve. This can be achieved by a conventional Quick-sort algorithm, assuming the availability of a comparison operator that compares points relative to their position on a z-curve. Efficient implementations of many operations on z-curves depend on properties of the representation of the curve position on bit-level, called Morton encoding [40]. We first present a short overview of this encoding based on two pixels. The Morton-encoded address of a position on the curve is called z-address. It consists of the interleaved bits of the two pixel values. In higher dimensions, the operation is performed analogously, such that the highest bit of each pixel is used before returning to the 2nd highest bit of the first pixel and so on.

An interesting observation is that the z-curve does not treat all dimensions equally. It rather introduces the concept of "more significant" or "less significant" dimensions. This is expressed by the order in which the different dimensions are processed when forming the z-address. In the two-dimensional example, it is possible to either use the x-address first, then the y-address, or the other way around. In fact, the two options result in two different z-curves. In the $n$-dimensional case, this leads to $n!$ possible curves. It is known from database research that indices give best performance if the curve uses dimensions that have the highest variability in a given dataset first. In our case, the dataset is generated using a PCA, so this property is guaranteed. Interestingly, the use of a PCA prior to building the z-curve also establishes rotational invariance of the method as any permutation of dimensions is removed by the PCA. Comparing two entries can be implemented efficiently without computing the entire z-address by calculating the most significant bit that differs between the two entries. Only the bits of this dimension have to be compared. The following paragraph shows a C++ code fragment to perform this operation.

```
inline bool lessMostSignificantBit(byte a, byte b)
{
    return ( a < b ) && ( a < (a ^ b) );
}
bool lessRelativeToZCurve( const byte* a, const byte* b )
{
    unsigned int bitWiseDifferenceSoFar = 0;
    unsigned int bestDimensionSoFar = 0;
    unsigned int size = (unsigned int) a.size();
    for (unsigned int dim = 0; dim < size; dim++)
    {
        auto bitWiseDifference = a[dim] ^ b[dim];
        if (lessMostSignificantBit ( bitWiseDifferenceSoFar,
            bitWiseDifference ) )
        {
            bitWiseDifferenceSoFar = bitWiseDifference;
            bestDimensionSoFar = dimension;
        }
    }
    return a[bestDimensionSoFar] < b[bestDimensionSoFar];
}
```

### *D. K-NN-Queries*

We now explain how to use this structure to perform a $k$-NN-search in the dictionary. We use a traversal technique called outside-in search [41]. Hereby, the acceleration structure is split into recursive regions, which are traversed depth-first. In each traversal step, the current region is split into two sub-regions, then the distance of both regions to the query object is computed. The sub-region closest to the query object is recursively traversed first, the sub-region further from the query object is traversed second. Hereby, the visit to the second sub-region is pruned if its distance to the query object is higher than that of the $k$-closest neighbors found so far.

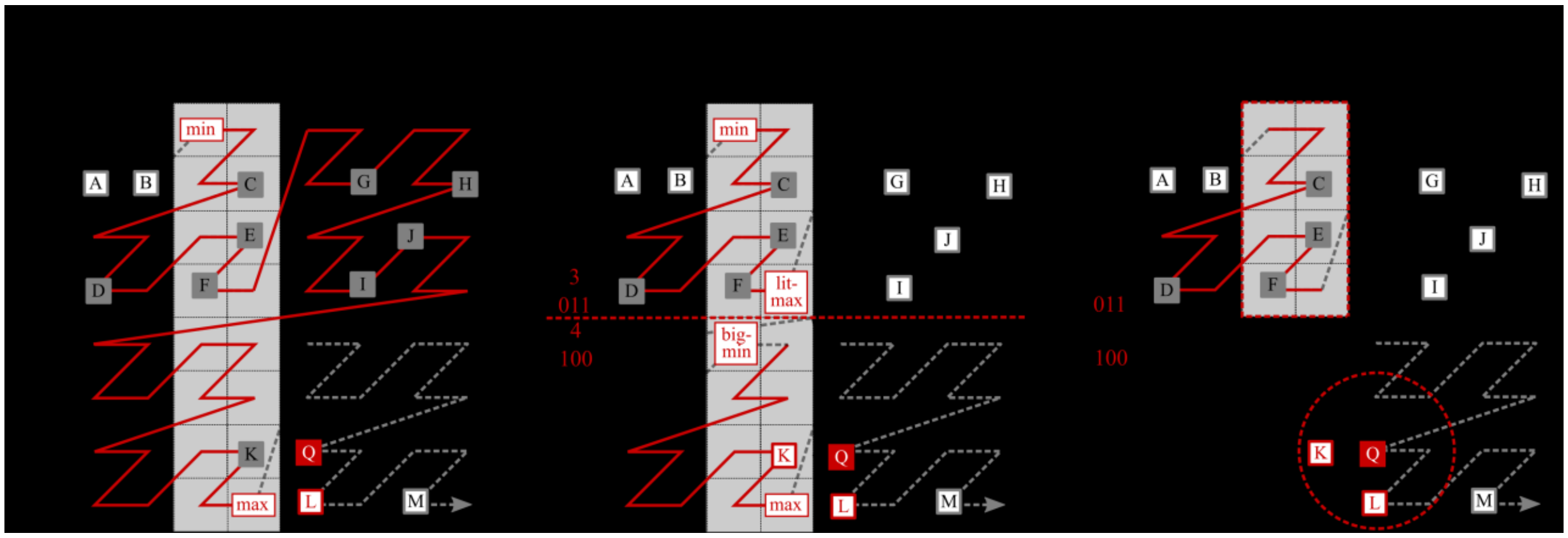

Figure 3. A *k*-NN -query on the z-curve index structure. The active search region is specified by an axis-aligned bounding box (min=2/0, max=3/7), indicated in gray. The query object at 4/6 is marked solid red, patches in the *k*-NN list are marked white/red. a) The example consists of 14 patches (A-M), sorted according to their position on the z-curve. One property of the z-curve is that all patches inside a bounding box are also inside the interval (min-max) on the z-curve. Those active patches are marked in gray, patches outside the active interval are marked white. b) The active region is split at litmax=3/3 and bigmin=2/4. The unification of the two resulting intervals (min-litmax) and (bigmin-max) does not cover the full interval (min-max), so patches G-J are culled. The two sub-regions are traversed according to their distance to the query object. c) After visiting the sub-region (bigmin-max), the distance of the query object to the remaining sub-region is compared to distance of the *k*-nearest neighbor found so far. The distance is higher, so the sub-region is not visited and patches C-F are culled.

In detail, the search operates by recursively subdividing the search space into axis-aligned regions. One property of the z-curve is that for each axis-aligned box specified by its upper left corner *min* and its lower right corner *max*, it holds that all points inside the box are also inside the interval (*min*,*max*) on the z-curve. With each axis-aligned region we can associate an interval on the z-curve (Figure 3a). The opposite is not true, so that in general the interval (*min*,*max*) contains some points outside the associated axis-aligned region. When recursively splitting the search regions, the goal is to minimize this overhead, i.e. find split positions where the unification of the intervals associated with the two sub-regions contains significantly less patches than the interval associated with the full region. This is achieved by the *litmax*/*bigmin* algorithm first presented in [32]. In short, the algorithm always splits the region at the position and in the dimension where most significant bits differ between the coordinates of *min* and *max*. Intuitively, this can be understood by inspecting the query region in Figure 3b. It can be observed that taking a step of size one at the position where the most significant bit of the coordinate changes corresponds to a large path along the curve. A proof that this scheme is in fact optimal is given in [33].

After the position of the split is determined, the algorithm computes the z-curve intervals associated with the two sub-regions. This is done by using a binary search on the dictionary to find the positions of *litmax* and *bigmin* on the z-curve. The intervals are then (*min*,*litmax*) and (*bigmin*,*max*). The two sub-regions are traversed recursively, whereby the order of traversal depends on the distance of the sub-region to the query object. The sub-region with the smaller distance is traversed first. After returning from the visit to the near sub-region, the algorithm checks if the far sub-region is farther away than the *k*-nearest neighbors found so far. If this is the case, the sub-region cannot contain a part of the result and the interval is culled (Figure 3c). At the lower levels of recursion, the sizes of the intervals fall below the defined threshold $\mu$. All patches contained in these intervals are checked using a full scan. For each point in these intervals, the distance of the point to the query-object is computed. If it is smaller than the distance to the *k*-nearest neighbors so far, the corresponding patch is added to the *k*-NN list. If the list overflows (contains more than *k* entries), the entry with the largest distance is removed from the list.

### *E. Performance Optimization*

While the described approach is straightforward, it turns out that a naïve implementation of the algorithm is quite slow. In order to optimize performance, we avoid storing a separate copy of *litmax* and *bigmin* on the stack for each recursion step and store only the dimension and position of the split. We can then re-use the memory already reserved for the query range, so that only one byte in the structure must be changed. A second optimization concerns the access to the dictionary. A binary search on a sorted vector of length $L$ is typically initialized from 0 to $L$-1 and then uses bisection to find an element. For the query algorithm, however, we already know an interval that must contain the searched element from the previous level of recursion. Therefore, it is wasteful to initialize the search from 0 to $L$-1 every time. Instead, the search can be initialized from the beginning to the end of the already computed interval. This interval can be substantially smaller than the dictionary, particularly on deeper levels of recursion, which account for the majority of query steps. In the following, we give C++ code for the recursive query algorithm including the discussed optimizations.

```
bool knnSearch(const Point& queryObject, HyperCube& activeRange,
                Interval interval, DataSet* dataset )
{
    if ( interval.last >= dataset->size () || interval.last < interval.first )
        return false;
    if ( interval.length() <= minimumIntervalLengthToUseRecursion )
        return fullScanSearch( queryObject, activeRange, interval, dataset);
    int splitDim = activeRange.dimensionWhereHighestBitDiffers();
    int splitPos = activeRange.positionWhereHighestBitDiffers();
    float leftDist = distanceToLeftRange( queryObject, activeRange, splitDim, splitPos );
    float rightDist = distanceToRightRange( queryObject, activeRange, splitDim, splitPos );
    if ( rightDist >= leftDist )
        return traverseLeftRight(queryObject, activeRange, interval, dataset, splitPos, splitDim, rightDist );
    else
        return traverseRightLeft (queryObject, activeRange, interval, dataset, splitPos, splitDim, leftDist );
}
```

The following listing gives code for the recursive routine that visits the sub-ranges in left-right order. Code for the right-left traversal works analogously.

```
bool traverseLeftRight(  const Point& queryObject,
                          HyperCube& activeRange,
                          Interval interval, DataSet* dataset )
{
    const byte maxInDimension = activeRange.last[splitDimension];
    const byte minInDimension = activeRange.first[splitDimension];
    activeRange.last[splitDimension] = splitPosition;
    Interval lower = dataset->dataPointsInInterval( activeRange, interval, KNOW_LOWER );
    bool didImprove = knnSearch( queryObject, activeRange, lower, dataset );
    activeRange.last[splitDimension] = maxInDimension;
    if ( rightDistance >= distanceToKthEntry() )
        return didImprove;
    activeRange.first[splitDimension] = splitPosition + 1;
    auto borders = ( didImproveDistance ) ? KNOW_NONE : KNOW_UPPER;
    Interval upper = dataset->dataPointsInInterval( activeRange, interval, borders );
    didImprove |= knnSearch( queryObject, activeRange, upper, dataset );
    activeRange.first[splitDimension] = minInDimension;
    return didImprove;
}
```

The next code segment is used at the lower levels of the recursion, as soon as the number of patches in the active interval falls below the threshold $\mu$. Whenever the result is improved, i.e. when a new entry is added to the $k$-NN-list, the active range is cropped. As the active range is passed by reference, this also reduces the active range on the previous (higher) levels of recursion.

```
bool fullScanSearch ( const Point& queryObject, HyperCube& activeRange,|
                      Interval interval, DataSet* dataset)
{
    bool didImprove = false;
    for (int i = interval.first; i <= interval.last; i++)
    {
        Point& dataPoint = dataset->getDataPointAtIntex(i);
        float distance = norm->distance( queryObject, dataPoint );
        if (distance < kDistanceEntry())
        {
            addToKnnList( KNNCandidate( distance, dataPoint.key ) );
            didImprove = true;
        }
    }
    if ( didImprove )
    {
        HyperCube newBoundingRange( queryObject, queryObject );
        newBoundingRange.extend( (int) std::ceilf( distanceToKthEntry() ) );
        activeRange.crop( newBoundingRange );
    }
    return didImprove;
}
```

### *F. Parallelization*

The traversal code given above can be parallelized to run efficiently on multi-core processors. However, a naïve implementation results in poor performance, as the subdivision into regions gives a strongly unbalanced partitioning of the search space. In order to achieve a reasonable load balancing, a fine granular parallelization is used that searches two sub-regions in parallel only if both regions are above a minimum size. This fine-grained subdivision means, however, that the overhead to starting a thread per sub-region makes it prohibitively slow. The problem can be solved by applying a thread pool design pattern [42]. Different from the original design pattern, we implemented a prioritized job queue where the distance between the query object and the traversed sub-region is used as a priority. This means, closer regions are always traversed first, independent of the order they were inserted into the queue. This increases the chance of pruning regions and improves performance.

The remaining issue is that access to the *k*-NN-list needs to be synchronized between threads. As read access to this list for finding the current relevant search radius is much more frequent than write access, which only occurs when a new candidate is inserted into the list, the synchronization can be solved efficiently using the read-write lock design pattern [43].

### *G. Acceleration with PatchMatch*

We compare the z-curve acceleration structure against PatchMatch [22], which is the current state-of-the-art for exemplar-based inpainting problems. PatchMatch is an acceleration structure applicable to various image processing problems including exemplar-based inpainting. The core idea of PatchMatch is the concept of propagating a nearest neighbor field (NNF). This means, spatial cohesion is exploited in the sense that if a patch at position $x$ has good match in the dictionary at position NNF($x$), then the patch at position $x+y$ will have a good match at NNF($x+y$) for small offsets $y$.

In order to allow an unbiased comparison, we adapt PatchMatch to the inpainting approach of Criminisi et al. [5], which requires some modifications to the PatchMatch algorithm. In particular, we implement PatchMatch in the same Criminisi framework as the z-curve accelerated inpainting and only replace the acceleration structure. PatchMatch assumes a fixed scanning order, while the Criminisi method is structure adaptive and thus each inpainted patch modifies the order in which the following patches are visited. We thus only replaced the part of the algorithm that selects the best patch in the dictionary, while the inpainting code and the computation of patch distance (i.e. the cost function) are identical to the z-curve variant of the algorithm. For more details on the implementation of the PatchMatch acceleration structure, see Appendix II.

### *H. Experimental Environment*

We implemented the algorithm described above in the C++ 11 programming language. The program was compiled with the native compiler of Microsoft Visual Studio 2017 using the AVX instruction set (/arch:AVX), fast floating point arithmetic (/fp:fast), and full compiler optimization (/Ox). Computation of the PCA was implemented based on the Intel DAAL library, which in turn links against the BLAS and LAPACK implementations provided by the Intel MKL library.

The final runtime experiments were conducted on an Intel Core i7-6850K at 3.6GHz with 96 GB memory. The brute-force (i.e. full scan from the beginning) experiments on a graphics processing unit (GPU) were performed on an NVidia GTX 1080Ti programmed using the OpenCL API. Z-curve parameter tuning experiments were conducted on an Intel Core i7-6700K at 4GHz with 16 GB memory.

### *I. Dataset for Evaluation*

We used a set of color images with resolutions of 800×600, 1600×1200, 2048×1536, and 2560×1920 pixels to evaluate our approach. For each size, a set of 10 different images was selected from a publicly available image database [44]. The evaluation data is provided as supplement data S2. Each image was deteriorated with a mask of text covering 20%±0.1% of the pixels (Figure 4).

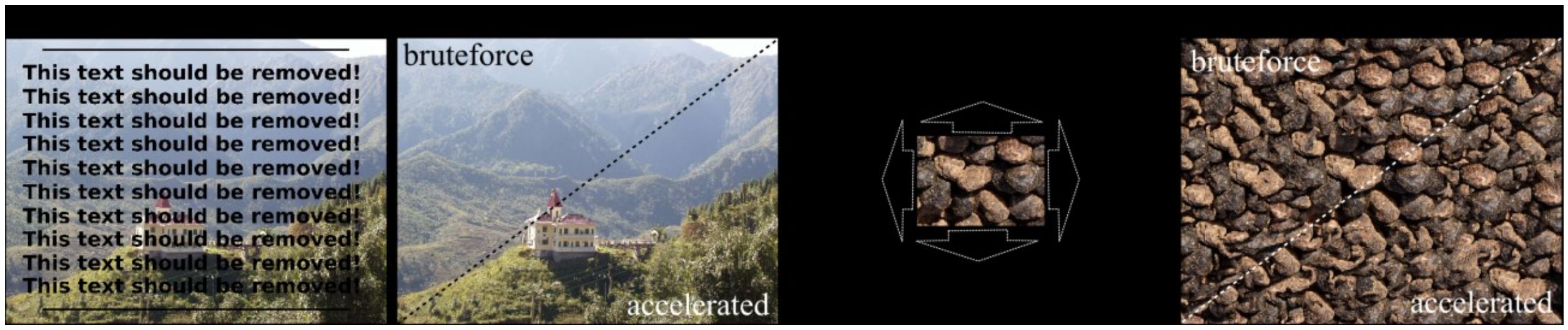

Figure 4. a) Example from the dataset used for evaluation of the results. a) Example with overlaid mask used to deteriorate the image. The mask consists of 80% white pixels and 20% black pixels. c) Result of inpainting with the GPU optimized brute-force versus the accelerated z-curve inpainting. c) Example of texture synthesis based on exemplar-based inpainting. The image is extended in all directions. d) Result of inpainting with the GPU optimized brute-force versus the accelerated z-curve inpainting.

## III. Results

The performance of the presented acceleration structure for exemplar-based inpainting depends on several parameters. Before gathering overall results we analyzed quality and performance parameters. We then selected values that constitute a good trade-off for the evaluated image resolutions between both satisfying results of the inpainting quality and fast runtimes.

### *A. Overall Performance*

We compared running times for several inpainting problems (Table 2). As a baseline, we used a highly optimized implementation of the brute-force search on a GPU and the original algorithm by Criminisi et. al. All experiments were conducted using the following parameters: $K$=9×9, $D$=10, $k$=80, $\mu$=256, $\nu$=2,048, and $c$=0.6. Explanations of how these parameters were obtained are given below.

*Table 2. Overall runtime of the algorithm with and without acceleration structure. Depicted are the average runtimes for ten images of each resolution ± the standard deviation of the runtimes. The acceleration structure runtimes include the time needed for dictionary learning. (s = seconds, m = minutes, h = hours).*

| dataset resolution | original CPU | brute-force GPU | z-curve CPU | PatchMatch CPU |
|---|---|---|---|---|
| 800×600 | 19.0±3.7 m | 29.5±0.3 s | 16.9±2.3 s | 46.1±9.0 s |
| 1,600×1,200 | 5.0±0.2 m | 2.7±0.04 m | 1.2±0.1m | 2.2±0.02 m |
| 2,048×1,536 | 12.6±1.1 h | 5.4±0.04 m | 1.8±0.2 m | 2.88±0.5 m |
| 2,560×1,920 | 35.4±2.2 h | 11.1±0.2 m | 3.2±0.3 m | 3.6±0.6 m |

### B. Sorting the Dictionary

The runtime of the acceleration structure also depends on the time it takes to sort the dictionary. We therefore evaluated this sorting time based on all 40 images contained in the evaluation dataset (Table 4). For the standard application, where damaged parts of an image are inpainted with the available parts of the image, the index structures need to be built prior to the inpainting procedure. However, there are also applications like sparse scanning microscopy inpainting, where a generic dictionary is used, so that index structures can be precomputed and stored on disc, which saves a lot of time.

### C. Impact on Inpainting Quality

The proposed multi-index structure as well as PatchMatch compute the best match for a patch approximately. We investigated the impact of this approximation on the quality of the inpainting by computing the $L_2$-norm of both z-curve/PatchMatch patch and brute-force patch for each iteration. The acceleration error (AE) is then defined as the z-curve/PatchMatch norm divided by the brute-force norm. For normalization, one is subtracted such that a value of zero corresponds to no acceleration error.

### D. Number of Dimensions of Principal Space D and Size of Candidate Set k

Two parameters determine the quality of the inpainting algorithm: the number of kept principal components $D$ and the number of entries $k$ in the nearest neighbors list. We evaluated values from $D$=4 to $D$=16 and from $k$=1 to $k$=320 on an image of 1,600×1,200 pixels resolution. The resulting acceleration error is displayed in Table 3. The influence of $D$ and $k$ on the runtime of the inpainting is displayed in Table 4.

*Table 3. Acceleration error (AE) for each combination of D and k. Combinations of D and k that result in an AE of <1% are considered acceptable and are marked by the black line (lower right corner). The combination D=10 and k=80 has the best runtime performance of the acceptable combinations and is indicated in bold letters (see also Table 4).*

| | **D** | | | | | | |
|---|---|---|---|---|---|---|---|
| **k** | **4** | **6** | **8** | **10** | **12** | **14** | **16** |
| **1** | 72.37 | 43.93 | 31.86 | 25.58 | 23.19 | 21.57 | 20.98 |
| **10** | 32.54 | 13.82 | 7.53 | 5.54 | 4.64 | 4.27 | 4.12 |
| **20** | 25.10 | 9.13 | 4.60 | 3.14 | 2.64 | 2.48 | 2.29 |
| **40** | 19.00 | 5.77 | 2.66 | 1.77 | 1.47 | 1.37 | 1.26 |
| **80** | 14.69 | 3.58 | 1.49 | **0.95** | 0.85 | 0.58 | 0.70 |
| **160** | 9.74 | 2.03 | 0.82 | 0.50 | 0.44 | 0.41 | 0.39 |
| **320** | 7.39 | 1.08 | 0.41 | 0.23 | 0.25 | 0.26 | 0.22 |

*Table 4. Runtime in seconds for each combination of D and k. Combinations of D and k that result in an AE of <1% are considered acceptable and are marked by the black line (lower right corner). Additionally, the average time for learning the dictionary is given as separate row. The combination with the best runtime performance and acceptable AE is indicated in bold letters.*

| | **D** | | | | | | |
|---|---|---|---|---|---|---|---|
| **k** | **4** | **6** | **8** | **10** | **12** | **14** | **16** |
| **1** | 11.26 | 12.21 | 13.96 | 16.23 | 18.78 | 21.82 | 25.45 |
| **10** | 12.27 | 13.83 | 16.15 | 19.30 | 22.15 | 25.94 | 30.00 |
| **20** | 13.53 | 15.11 | 17.65 | 21.24 | 24.40 | 28.44 | 32.94 |
| **40** | 15.64 | 17.72 | 20.12 | 24.11 | 28.39 | 32.83 | 37.91 |
| **80** | 21.15 | 23.85 | 27.06 | **31.66** | 36.41 | 41.24 | 46.45 |
| **160** | 35.60 | 40.30 | 44.20 | 50.00 | 55.35 | 61.05 | 66.13 |
| **320** | 65.73 | 73.46 | 79.90 | 90.09 | 98.43 | 105.57 | 112.25 |
| **sorting** | 6.7 | 7.4 | 7.7 | 8.3 | 8.3 | 8.6 | 8.7 |

We selected $D$=10 and $k$=80 for all further experiments, as it represents a reasonable trade-off between quality and performance. This combination had the best runtime among all evaluated combinations with AE<1%.

### *E. Optimal Recursion Thresholds μ*

The parameter $\mu$ controls when the recursion stops. The algorithm then processes the remaining patches in the dataset sequentially. We determined an optimal value by running inpainting experiments on one image for each resolution for values of $\mu$ ranging from 16 to 4,096 (Table 5). We found that the best choice for the recursion threshold enhances running times by up to 40%. The optimal value was consistently at $\mu$=256 for any of the evaluated images, so that this value was selected for all further experiments.

*Table 5. Influence of the recursion threshold μ on the runtime of the accelerated inpainting algorithm in seconds. For each resolution, a single image was processed.*

| | **Resolution [pixels]** | | | |
|---|---|---|---|---|
| ***μ*** | **800×600** | **1,600×1,200** | **2,048×1,536** | **2,560×1,920** |
| 16 | 9.9 | 38.2 | 70.0 | 135.5 |
| 32 | 8.6 | 34.1 | 63.1 | 115.2 |
| 64 | 7.7 | 31.1 | 58.2 | 105.3 |
| 128 | 7.2 | 29.2 | 56.0 | 100.1 |
| 256 | **7.1** | **28.5** | **55.2** | **97.9** |
| 512 | 7.2 | 29.0 | 56.3 | 98.1 |
| 1024 | 7.4 | 30.7 | 58.6 | 99.0 |
| 2048 | 7.9 | 32.8 | 62.0 | 104.2 |
| 4096 | 8.4 | 35.2 | 66.1 | 110.5 |

### *F. Optimal Parallelization Thresholds v*

Another important performance parameter is the threshold for parallelization $v$. We determined the value by running inpainting experiments on one image for each resolution for values of $v$ ranging from 16 to 8192. An optimal choice of the parallelization threshold can also speed up the computation by up to 50%. Optimal values for $v$ vary between 512 and 4096 (Table 6). We selected $v$=2048 for all further experiments, as this value gave fastest running times for most resolutions.

*Table 6. Influence of the parallelization threshold v on the runtime of the accelerated inpainting algorithm in seconds. For each resolution, a single image was processed.*

| | **Resolution [pixels]** | | | |
|---|---|---|---|---|
| ***v*** | **800×600** | **1,600×1,200** | **2,048×1,536** | **2,560×1,920** |
| 16 | 10.5 | 52.7 | 111.6 | 159.2 |
| 32 | 9.9 | 49.7 | 104.0 | 153.8 |
| 64 | 9.1 | 46.4 | 95.8 | 139.2 |
| 128 | 8.0 | 40.4 | 80.9 | 125.5 |
| 256 | 7.3 | 35.0 | 67.4 | 105.6 |
| 512 | **7.0** | 30.7 | 59.3 | 91.7 |
| 1,024 | 7.1 | 28.9 | 56.4 | 85.1 |
| 2,048 | 7.1 | **28.4** | **55.6** | 82.3 |
| 4,096 | 7.4 | 28.9 | 56.7 | **82.2** |
| 8,192 | 7.6 | 31.0 | 58.3 | 85.7 |

### *G. Optimal Value c*

The last performance parameter we determined was the percentage of pixels that are covered by the applied indices. We determined the value for *c* by running an experiment based on a 2,560×1,920 resolution image for values between 0.2 and 0.8. Runtimes increased monotonically from 120 seconds for $c$=0.2 to 263 seconds for $c$=0.8. As values below $c$=0.6 created clearly visible artifacts we decided to set *c* to 0.6. Depending on the missing parts in an image that should be inpainted, *c* might be set to lower values for faster runtimes without much quality loss, e.g. for clear textures or homogeneous areas.

### *H. Results for iDistance Index*

Additional to the z-curve acceleration structure, we investigated the use of iDistance, which is one of the state-of-the-art multi-dimensional indexing structures. It is used for *k*-NN queries especially when the dimension of the data is very high. We tested iDistance on an image with a resolution of 1,600×1,200 pixels but found that the runtime performance was clearly inferior compared to the optimized GPU accelerated implementation. Because of this result, we abandoned the idea to use iDistance as multi-dimensional indexing structure and decided to use the *k*-NN search on the space-filling curves, which turned out to work well as already shown.

### *I. Results for PatchMatch*

We compare our algorithm against PatchMatch as the current state-of-the-art for acceleration of patch-based algorithms. Our experiments reveal that the efficiency of PatchMatch in its original form [22] can be diminished if it is adapted to priority-based algorithms such as the one of Criminisi et al. The nearest neighbor field for known pixels adjacent to the fillfront has to be initialized with a PatchMatch preprocessing step before the actual inpainting. This generates a significant runtime overhead, particularly for low resolutions (see Appendix II for more detailed explanation). This also explains why PatchMatch on the CPU falls behind brute-force on the GPU in the 800x600 resolution in Table 2, but outperforms it for higher resolutions. In general, PatchMatch as an acceleration structure performs worse than our proposed z-curve acceleration for all image sizes, but becomes more competitive for higher resolutions. Nevertheless, PatchMatch performs well in terms of quality (compare Figure 5), which is also reflected by an average AE of 0.61±0.40.

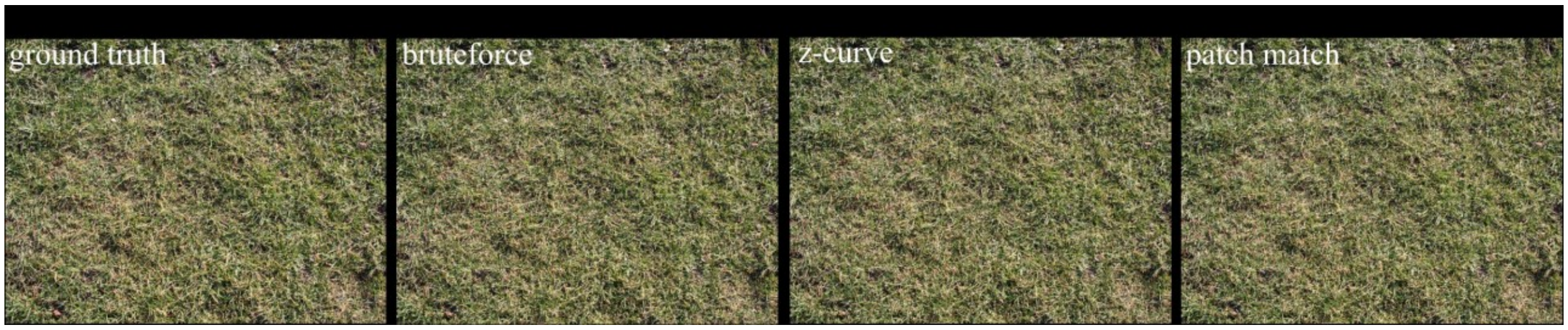

*Figure 5. Comparison of the implemented approaches against the unaltered (ground truth) image.*

### *J. Theoretical Performance*

For a constant number of dimensions of the principal space, the algorithm has a theoretic runtime of $O(\log^2 n)$, where $n$ is the size of the dictionary. A proof of this result is provided as supplement S1. This result holds with very mild assumptions on the cost function, it is sufficient that the cost function is a norm. Furthermore, we obtain that the algorithm scales with $O(\sqrt[D]{(D/2)!})$ in the number of dimensions of the principal space. It therefore seems advisable to select a moderate number of dimensions for the principal space, which is consistent with our experimental results (Table 4).

### *K. Performance for $L_1$-norm as Cost Function*

We experimentally investigated the performance of the acceleration structure if the $L_2$-norm is replaced by the $L_1$-norm. As expected, the acceleration settings had to be adjusted in order to keep the acceleration error in the acceptable range. We changed the values from $D$=10 and $k$=80 (Table 3) to $D$=14 and $k$=160. Using these settings, we obtained an acceleration error of approximately 1%, and the inpainting took 62.7 seconds (plus 8.0 seconds for building the indices) for our test image. Thus, the acceleration structure is approximately two times slower when using the $L_1$-norm compared to the $L_2$-norm.

## IV. Discussion & Conclusions

We presented an acceleration structure for exemplar-based inpainting that uses concepts originally developed in the field of high-dimensional indexing, for example in multimedia databases. From a theoretical point of view, the acceleration structure has runtime in $O(\log^2 n)$ per query, where $n$ is the size of the dictionary. This is a fundamental improvement over the original inpainting proposed by Criminisi et al., which has a runtime in $O(n)$ per query.

In terms of practical performance gains, implementation, and dictionary size play an important role. As a baseline, we implemented a highly optimized version of the original algorithm that runs in parallel on a GPU. Compared to this baseline, our method achieved a speedup of a factor of 1.5 to 3.5 depending on the size of the inpainted images and hence also the dictionary size. Compared to the original algorithm, we obtained an average speedup factor of 236 on the largest images. Also, our method was 1.2 to 3 times faster than acceleration based on the current state-of-the-art PatchMatch. The real importance of the acceleration structure lies in the logarithmic scalability. Using the proposed method, extending exemplar-based inpainting from two-dimensional images to three-dimensional datasets with much larger dictionaries becomes a feasible option. This will allow for novel applications, including exemplar-based inpainting of video data and exemplar-based inpainting of volumetric data, for example in three-dimensional microscopy. A video showing a time-series of exemplar-based inpainting as a texture synthesis algorithm is provided as supplement S3.

An interesting aspect of the acceleration structure is that the inpainting requires the construction of several index structures as a preprocessing step. For conventional inpainting, the undamaged parts of the image are used as dictionary. In such scenarios, the index structures need to be built for each inpainting as it was the case for the performance evaluation given in Table 2. For other applications, for example in sparse microscopy, it is possible to reuse dictionaries, which opens a potential for additional performance improvements. On the downside, the need to initialize the indices means that the proposed acceleration structure is virtually useless for cases where only few pixels need to be inpainted, as the cost for the initialization cannot be amortized over a too small number of query operations.

The proposed acceleration structure is independent of the exact inpainting algorithm in two ways. First, the structure is used to accelerate individual queries. As such, it can be used in combination with any exemplar-based inpainting algorithm, independent of the heuristic that defines the order in which patches are processed. Second, the theoretical result of logarithmic runtime was obtained with very mild assumptions about the cost function. Thus, the acceleration structure can be used with any norm as cost function, at least in theory. In practice, we found that the computational cost of evaluating a norm plays an important role. For example, a general $L_p$-norm requires the evaluation of logarithmic as well as exponential functions and is non-trivial to implement efficiently.

The main contribution lies on the fact that the acceleration structure is independent from most model-based aspects of the inpainting algorithm. Thus, it can be combined with a large number of existing and future exemplar-based inpainting algorithms. The acceleration structure makes the generalization of exemplar-based inpainting to three-dimensional datasets such as video or volumetric data a feasible option.

## V. Additional Information

### *A. Acknowledgements*

The authors thank the DFKI GmbH and Saarland University for funding the research and for providing the necessary infrastructure. This project has received funding from the European Research Council (ERC) under the European Union's Horizon 2020 research and innovation program (grant agreement no. 741215, ERC Advanced Grant INCOVID).

### *B. Competing financial interests*

There are no competing financial interests.